\pdfoutput=1
\documentclass[10pt]{article}

\usepackage[accepted]{tmlr}

\usepackage{amsmath,amssymb}
\usepackage{graphicx}
\usepackage{booktabs}
\usepackage{xcolor}
\usepackage{placeins}
\usepackage{float}
\usepackage{array}
\usepackage{tabularx}
\usepackage{url}
\usepackage[colorlinks=true,linkcolor=blue,citecolor=blue,urlcolor=blue,bookmarks=false,
  pdftitle={Metamorphic Testing of Learned Spatial Surrogates for Transit Accessibility},
  pdfauthor={Till Zacher and Markus Lienkamp}]{hyperref}

\newcommand{\doi}[1]{doi: \begingroup\urlstyle{rm}%
  \href{https://doi.org/#1}{\nolinkurl{#1}}\endgroup}

\newcommand{\scorename}[1]{\texttt{#1}}
\newcommand{\isl}{\scorename{isl}}
\newcommand{\prs}{\scorename{prs}}
\newcommand{\rcs}{\scorename{rcs}}
\newcommand{\wal}{\scorename{wal}}
\newcommand{\odr}{\scorename{odr}}
\newcommand{\sdl}{\scorename{sdl}}
\newcommand{\bis}{\scorename{bis}}
\newcommand{\sel}{\scorename{sel}}

\newcommand{\ufone}{U-Net}
\newcommand{\raone}{ResU-Net}
\newcommand{\dstwo}{Dual-stream}
\newcommand{\tsthree}{Triple-stream}

\newcommand{\bus}{\mathbf{B}}
\newcommand{\rail}{\mathbf{R}}
\newcommand{\poi}{\mathbf{P}}
\newcommand{\tdelta}{\boldsymbol{\delta}}

\newcommand{\model}{f_\theta}

\def\month{11}
\def\year{2026}
\def\openreview{\url{https://openreview.net/forum?id=bG1yIXRC8Y}}

\title{Metamorphic Testing of Learned Spatial Surrogates for Transit Accessibility%
  \thanks{The authors used OpenAI ChatGPT 5 as an assistive tool for language editing and code support. All scientific ideas, claims, experimental decisions,
  and results are human-sourced. The authors verified all citations and final content and take full
  responsibility for the manuscript.}%
}

\author{%
  \name Till Zacher \email till.zacher@tum.de \\
  \addr Institute of Automotive Technology, Technical University of Munich \\
  Munich Institute of Robotics and Machine Intelligence (MIRMI) \\
  Garching, Germany
  \AND
  \name Markus Lienkamp \email lienkamp@tum.de \\
  \addr Institute of Automotive Technology, Technical University of Munich \\
  Munich Institute of Robotics and Machine Intelligence (MIRMI) \\
  Garching, Germany
}

\begin{document}

\maketitle

\begin{abstract}
Transit-network design requires evaluating many candidates, but detailed accessibility calculations
are too slow to repeat inside an optimization loop. We therefore train a neural surrogate to
approximate them. Given a candidate
bus network and fixed information about rail service and nearby opportunities, it predicts the
calculated spatial change in accessibility. We compare its predictions with routing-derived targets
in geographically held-out regions and test specified qualitative relations under controlled
input changes. Reference calculations on constructed networks assess the applicability of
relations that have a routing-level counterpart.
Models with positive predictive skill can still fail basic connectivity tests.
One targeted intervention substantially increases pass rates for two connectivity relations while
predictive skill also improves. Average prediction error is therefore insufficient when a learned model becomes
an optimization objective. Predictive agreement and controlled behavioral tests should be reported
separately.

\end{abstract}

\section{Introduction}
\label{sec:intro}

Accessibility, the ease with which people can reach spatially distributed
opportunities, is a long-standing objective and evaluation concept in transport and land-use
planning \citep{hansen1959accessibility}. A common operationalization is cumulative-opportunity
accessibility, which aggregates opportunities reachable within a specified travel-time threshold
\citep{tomasiello2023timeinterval}. Here, accessibility weights opportunities by an exponential
function of travel time within a cutoff. Evaluating candidate bus networks requires multimodal
routing and opportunity aggregation, often too costly to repeat throughout a design search.

\begin{figure}[t]
  \centering
  \includegraphics[width=\textwidth]{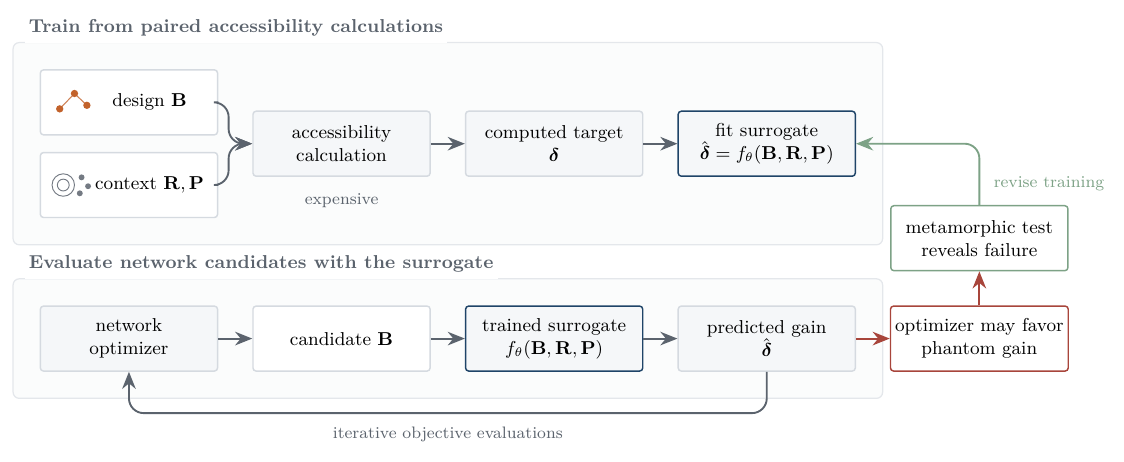}
  \caption{%
    \textbf{From reference accessibility calculations to a fast design objective.}
    Paired calculations train a surrogate that can score candidate networks during
    optimization. Metamorphic tests expose violations of specified structural relations.
  }
  \label{fig:surrogate_story}
\end{figure}

Our solution is to pay this cost offline. For each sampled bus network, a routing-based accessibility
calculation compares the same city with and without the candidate service. We call this the reference
calculation. Its computed differences supervise a spatial model that predicts accessibility change
from the candidate network and fixed rail and opportunity context. The trained surrogate then
provides a fast approximation intended for repeated evaluation inside an optimizer
(Figure~\ref{fig:surrogate_story}). It does not predict routes, demand, or observed ridership.

The reference calculation combines real OpenStreetMap, timetable, and opportunity data. A companion
study compares its travel-time impedance with smartphone travel diaries, without claiming complete
validation of the resulting accessibility fields \citep{zacher2026accessibility}. Here we ask whether
the surrogate reproduces computed targets in held-out regions and responds plausibly to controlled
changes. Those targets are counterfactual accessibility values, not observed travel behavior.

Putting the learned model inside an optimizer changes what counts as a good approximation.
Held-out error summarizes performance on passively sampled examples. An optimizer can instead
construct inputs that maximize an objective derived from the surrogate and preferentially seek
regions in which model error produces spuriously favorable values \citep{trabucco2021conservative}.
A disconnected stop assigned positive accessibility gain could therefore become attractive when
the predicted field is aggregated into a design objective. Average error cannot identify that
failure class on its own.

We compare four neural-network families that differ in how they combine the three inputs. U-Net
and residual U-Net encode the bus network, rail, and opportunities together. The dual-stream model
separates the bus design from the combined fixed context, while the triple-stream model also
separates rail from opportunities. This ordered comparison tests whether separating design from
context changes behavior and whether splitting the context further adds value. Capacity-matched
variants control for model size.

Our primary methodological contribution is an automated metamorphic testing protocol.
Controlled input changes test expected responses, such as lower accessibility when a stop is
disconnected. The tests complement prediction error by checking whether the model responds
consistently to changes that matter for network design.

Table~\ref{tab:design} maps the four empirical questions to the comparisons, controls, and
evidence used to answer them.

\begin{table}[t]
\centering
\footnotesize
\setlength{\tabcolsep}{3.2pt}
\renewcommand{\arraystretch}{1.15}
\caption{\textbf{Research questions and controlled comparisons.} The study varies one main axis
within each comparison rather than treating all architecture and training choices as a complete
factorial design.}
\label{tab:design}
\begin{tabularx}{\textwidth}{@{}>{\raggedright\arraybackslash}p{0.22\textwidth}
  >{\raggedright\arraybackslash}X >{\raggedright\arraybackslash}X
  >{\raggedright\arraybackslash}X@{}}
\toprule
Research question & Comparison & Held fixed & Evidence \\
\midrule
Can predictive skill hide structural failures? & Held-out geographic prediction versus generated
metamorphic tests & Same unaugmented model instances and output channels & Predictive skill and
eight probe pass rates \\
Can targeted data reduce a known failure? & Incremental training-data additions in the triple-stream
sequence, followed by a combined treatment in the dual-stream family & Architecture within each sequence and the common test
suite & Predictive skill and probe pass rates across three initializations \\
Does separating the bus input from context change model behavior? & Four input architectures,
from early fusion to separated streams & Training treatment, data split, evaluation suite, and
approximately matched capacity & Intensity saturation and seven companion probes \\
How does predictive agreement vary by opportunity category? & All 105 reconstructed output
channels of the selected dual-stream configuration & Three model instances and the same 818
geographically held-out tiles & Per-channel $R^2$ and target variation \\
\bottomrule
\end{tabularx}
\end{table}

Training on disconnected stops paired with zero accessibility-change targets improves
connectivity responses in two model families. Other interventions are unstable or ineffective,
and architecture affects some responses but not others. These findings motivate evaluating
design surrogates through both prediction error and controlled behavioral tests.

\section{Related Work}
\label{sec:related}

\paragraph{Spatial surrogates and heterogeneous inputs.}
Surrogate models replace expensive simulations with response surfaces that can be evaluated inside
an iterative design procedure \citep{queipo2005,jin2011}. For spatial outputs, encoder--decoder
networks such as U-Net preserve fine spatial detail while learning multi-scale features
\citep{ronneberger2015}; neural operators extend the same broad idea to mappings between functions
\citep{li2021,lu2020}. Geospatial and remote-sensing models often combine optical, radar,
elevation, and temporal modalities through early fusion, modality-specific encoders, or attention
\citep{huang2025,garioud2023,roy2023,wang2024a}. Our comparison asks whether separating a variable
bus design from fixed context is associated with better structural responses.

\paragraph{Evaluation under optimization pressure.}
When predictions enter a downstream decision, small average error need not imply good decisions.
Decision-focused and predict-then-optimize methods make this mismatch explicit by evaluating errors
through the optimization problem they induce \citep{wilder2019,elmachtoub2022}. Distribution shift
can also be endogenous: deployed predictions change which inputs are selected
\citep{perdomo2020}. More directly, offline model-based optimization and model-based control show
that optimization can concentrate on regions where a learned objective or dynamics model makes
systematically favorable errors \citep{trabucco2021conservative,eysenbach2022mismatched}.
Conventionally trained simulation surrogates can likewise misjudge intervention effects and direct
decision-makers toward inferior interventions \citep{dyer2024interventionally}. These perspectives
motivate pre-deployment behavioral testing as a complement to held-out prediction error.

\paragraph{Metamorphic testing.}
Metamorphic testing addresses problems for which exact output oracles are unavailable or expensive
by checking expected relations between related inputs \citep{segura2016,chen2019,chen2020}. It has
been applied to machine-learning systems using transformations that should preserve, order, or
otherwise constrain outputs \citep{xie2011,zhang2018}. Scientific regression studies provide a
closer precedent for our setting: conventionally accurate learned models can still violate
domain-expected responses to controlled perturbations \citep{yang2021reliability,reichert2024metamorphic}.
A complementary line of work uses discovered violations to train models toward specified
functional relations \citep{dreossi2018counterexample,sivaraman2020counterexample}. We apply these
ideas to spatial accessibility fields through generated, channelwise tests and targeted training
examples.

\paragraph{Transit accessibility.}
Open workflows that combine OpenStreetMap street data with GTFS schedules support reproducible
multimodal routing and accessibility analysis \citep{pereira2021r5r}. Accessibility has also been
optimized directly in transit-network design through cumulative-opportunity objectives under
operating constraints \citep{rayaprolu2022maximizing}, while learning-based heuristics have also
been studied for transit-network design \citep{holliday2025}. We examine whether accessibility
surrogates reproduce structural responses as well as held-out target fields.

\section{Methods}
\label{sec:methods}

\subsection{Prediction Task and Supervised Examples}
\label{sec:prob}

Each example is a spatially registered square spanning 21 km in EPSG:3857 coordinates, with height
$H=200$ pixels and width $W=200$ pixels, or 105 projected metres per cell. This geometry was chosen
to balance local route detail, surrounding context, and computational cost.

The candidate bus network is represented by stop-density and route-edge rasters,
$\bus\in\mathbb{R}^{2\times H\times W}$. Fixed rail-accessibility and opportunity-density
contexts each originate from the same 105-layer accessibility schema: 97 point-of-interest
categories and eight workplace layers. This inherited schema keeps destination types separate and
is not tuned. A principal component analysis (PCA) basis fitted to training-tile opportunity vectors reduces
each context to $K=32$ channels, retaining 83.7\% of variance in the PCA fitting sample.
This reduces dimensionality, memory use, and learning difficulty. The rank and basis are fixed
for every model, giving
$\rail,\poi\in\mathbb{R}^{K\times H\times W}$.

\begin{figure}[H]
  \centering
  \includegraphics[width=0.84\textwidth]{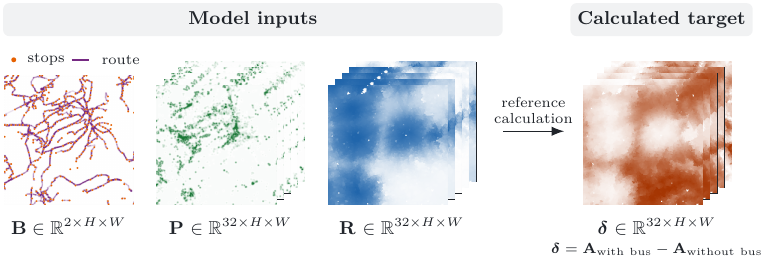}
  \caption{%
    \textbf{One routing-derived training tile.}
    The candidate bus design and fixed context predict the bus-induced accessibility change.
  }
  \label{fig:supervised_tile}
\end{figure}

For every sampled bus network, the reference calculation produces 105-channel accessibility fields
with and without that service under the same fixed city context. After projection into the retained
PCA basis, we denote these fields by
$\mathbf A_{\mathrm{with\ bus}},\mathbf A_{\mathrm{without\ bus}}
\in\mathbb{R}^{K\times H\times W}$. Their difference is the bus-induced response
\begin{equation}
  \tdelta=\mathbf A_{\mathrm{with\ bus}}-
  \mathbf A_{\mathrm{without\ bus}}.
  \label{eq:delta}
\end{equation}
Figure~\ref{fig:supervised_tile} shows one training example. Five aligned components
are displayed from each 32-channel PCA tensor and scaled independently to the 99.5th percentile of
their absolute coefficients. Training uses the complete signed tensors with fixed channelwise
normalization.

\begin{figure}[H]
  \centering
  \includegraphics[width=\textwidth]{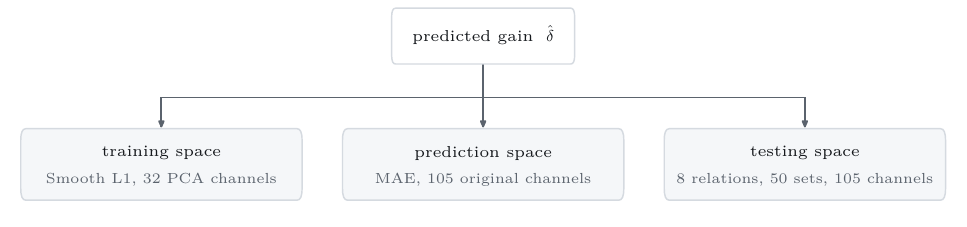}
  \caption{%
    \textbf{One prediction, three evaluation spaces.}
    Training loss, held-out target agreement, and structural testing remain separate.
  }
  \label{fig:evaluation_spaces}
\end{figure}

The surrogate emits a normalized 32-channel representation
$\hat{\mathbf z}=\model(\bus,\rail,\poi)$. With $\mathbf z$ the normalized form of
$\tdelta$, training minimizes pixelwise Smooth L1 loss,
\begin{equation}
  \mathcal L(\theta)=\frac{1}{KHW}\sum_{kij}
  \ell_{\mathrm{SL1}}(\hat z_{kij},z_{kij}).
  \label{eq:loss}
\end{equation}
Predictions are denormalized and inverse-projected to the original 105 accessibility channels for
both predictive and metamorphic evaluation. We refer to this original-channel representation as
raw space. Figure~\ref{fig:evaluation_spaces} keeps the three resulting spaces
separate. Training loss measures optimization in normalized PCA space, raw-space error measures
agreement with held-out routing-derived targets, and metamorphic pass rates measure responses to
controlled input changes.

\subsection{Data and Geographic Prediction}
\label{sec:data}

Tiles are generated from OpenStreetMap networks and General Transit Feed Specification (GTFS)
schedules, an established open-data pattern for multimodal accessibility analysis
\citep{pereira2021r5r}. Targets are differences between paired reference calculations.

All candidates share fixed operating assumptions, rail, and opportunity context on both sides of
Equation~\ref{eq:delta}, isolating bus-network topology. The bus channels represent stops and route
geometry, not a complete service plan; operational variables are controls rather than model inputs.

The geographic split contains 14,488 training tiles, 1,610 validation tiles, and 818 test tiles.
The test set withholds a 35-km-radius Munich catchment and a rural bounding box in
Mecklenburg-Vorpommern. A 25.5-km projected buffer excludes neighboring training footprints. PCA
and normalization statistics are fitted on training tiles only, and the split is fixed across
initialization seeds. This separation reduces the optimistic dependence that can arise when
spatially autocorrelated observations are randomly mixed between training and evaluation sets
\citep{roberts2017crossvalidation}. Pooled results use all 818 tiles; regional results use the
corresponding subsets. Both follow the same masked 105-channel protocol.

The zero predictor has MAE $2.997$; reconstruction of rank-32 projected targets has MAE $1.069$.
The latter is a reference, not a proven lower bound on prediction MAE. We define
\begin{equation}
  \operatorname{skill}=\frac{2.997-\operatorname{MAE}}{2.997-1.069},
  \label{eq:skill}
\end{equation}
so zero matches the zero predictor's MAE and one matches the PCA reconstruction MAE.
For the selected model family, we additionally pool masked pixels across all
test tiles and report each raw channel's coefficient of determination,
$R_c^2=1-\mathrm{SS}_{\mathrm{res},c}/\mathrm{SS}_{\mathrm{tot},c}$. We display every channel but
rank only those with target standard deviation of at least $0.2$ raw units, avoiding unstable
$R^2$ values when target variation approaches zero. This descriptive rule affects no other result.

All models use AdamW at initial learning rate $10^{-3}$ with cosine decay, at most 1,000 epochs,
and dropout 0.1. Each checkpoint contains the best-validation weights from one initialization. Appendix
\ref{app:training} gives the remaining training and split details.

\subsection{Automated Metamorphic Testing}
\label{sec:eval_framework}

The suite evaluates eight metamorphic relations across 50 independently generated scenario sets,
each containing reference, modified, and control scenes. Geometry and opportunity distributions
are parameterized from training data; every fifth set activates all 105 opportunity channels to
cover rare categories. Appendix~\ref{app:instrument} details generation and invariant checks.

\begin{figure}[t]
  \centering
  \includegraphics[width=0.68\textwidth]{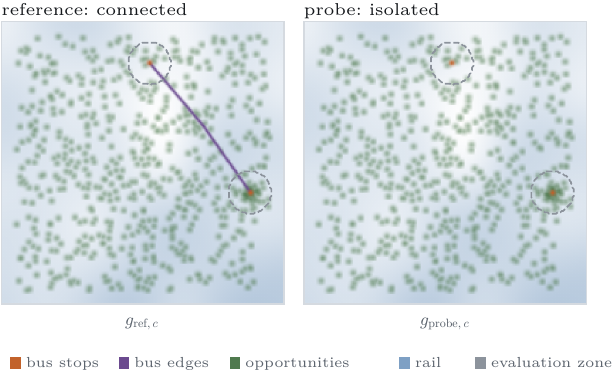}
  \caption{%
    \textbf{Anatomy of one generated relation.}
    A controlled reference--probe pair tests the expected ordering in each supported raw channel.
  }
  \label{fig:test_anatomy}
\end{figure}

The pair in Figure~\ref{fig:test_anatomy} holds the bus stops, opportunity context, rail context,
and evaluation zone constant. The probe removes bus edges while leaving stops and context visible.

Let $\hat\delta_c(s)$ be the reconstructed prediction for raw channel $c$ in scenario $s$. For a
stored evaluation mask $\mathcal M_s$, its non-negative masked gain is
\begin{equation}
  g_c(s)=\frac{1}{|\mathcal M_s|}\sum_{(i,j)\in\mathcal M_s}
  \max\!\left(\hat\delta_{cij}(s),0\right).
  \label{eq:channel_gain}
\end{equation}
Each relation supplies a reference gain $g_{\mathrm{ref},c}$ and probe gain
$g_{\mathrm{probe},c}$. Leakage scores use
$g_{\mathrm{probe},c}/(g_{\mathrm{ref},c}+g_{\mathrm{probe},c})$; sensitivity scores use
$g_{\mathrm{ref},c}/(g_{\mathrm{ref},c}+g_{\mathrm{probe},c})$; saturation scores use the probe's
share of their sum. Thresholds are operational conventions, not
physical constants. Leakage below $0.25$ and sensitivity above $0.75$ require a threefold separation
between the expected stronger and weaker responses. Saturation inside $[0.35,0.65]$ permits neither
response to exceed the other by more than approximately $1.86\times$. Thresholds apply to each
set--channel observation, not to configuration averages.

Opportunity displacement isolates the contribution of a focal opportunity from unchanged
background gain. Conditions $A$ and $B$ place it near and away from the route; a background-only
condition $N$ omits it. All three retain the same network, remaining context and evaluation mask:
\begin{equation}
 F_{A,c}=g_c(A)-g_c(N),\qquad F_{B,c}=g_c(B)-g_c(N),\qquad
 S_c=\frac{F_{A,c}}{F_{A,c}+|F_{B,c}|}.
 \label{eq:opportunity_effect}
\end{equation}
On channels containing the focal opportunity, the test requires $F_{A,c}\geq\tau_c$ and $S_c>.75$.
The absolute value penalizes negative residuals too, so attenuation does not establish monotonicity.
Appendix~\ref{app:opportunity_correction} reports the signed-response diagnostics and measurement details.

A set--channel observation is supported only when its reference gain reaches the fixed category-specific
floor
\begin{equation}
  \tau_c=\max\!\left(10^{-6},0.01\,q_{.25,c}^{+}\right),
  \label{eq:reference_floor}
\end{equation}
where $q_{.25,c}^{+}$ is the 25th percentile of positive training-tile mean target gain for
channel $c$. This training-derived floor excludes effectively inactive references and prevents
unstable ratios. For opportunity displacement it applies to $F_{A,c}$ rather than total gain.
Unsupported observations are excluded, not counted as passes; coverage and all-eligible passing
fractions accompany the conditional opportunity results. For model instance $m$ and relation $r$,
the pooled supported-observation pass rate is
\begin{equation}
  \hat p_{mr}=\frac{\sum_{s,c}v_{msrc}\,\mathbf{1}\{\text{relation passes}\}}
  {\sum_{s,c}v_{msrc}},
  \label{eq:pass_rate}
\end{equation}
where $v$ marks support. Uncertainty uses a deterministic 2,000-replicate percentile bootstrap
that resamples the 50 generated sets as clusters. Channels and pixels remain inside their sets;
they are not treated as independent replicates. Independently initialized checkpoints are the
replication unit for model-level claims.

\begin{table}[t]
\centering
\footnotesize
\setlength{\tabcolsep}{3.5pt}
\caption{\textbf{Generated probes and per-observation thresholds.} Reference checks distinguish
calculated counterparts, analytic zero-service controls, and raster-amplitude stress tests.}
\label{tab:relations}
\begin{tabularx}{\textwidth}{@{}>{\raggedright\arraybackslash}p{0.18\textwidth}lllXl@{}}
\toprule
Relation & Check & Kind & Reference $\rightarrow$ probe & Controlled change & Pass \\
\midrule
Isolated stop (\isl) & analytic & leakage & connected $\rightarrow$ isolated stop & bus edges removed & $<.25$ \\
Stub edge (\sel) & analytic & leakage & full route $\rightarrow$ endpoint stub & bus edge truncated & $<.25$ \\
Spatial localization (\sdl) & routing & leakage & route zone $\rightarrow$ equal-area far zone & evaluation region & $<.25$ \\
Opportunity reachability (\prs) & routing & attenuation & near $\rightarrow$ displaced focal & focal location; shared $N$ & $>.75$ \\
Rail-context sensitivity (\rcs) & analytic & sensitivity & connected $\rightarrow$ no bus edges & bus edges removed & $>.75$ \\
Weak activation (\wal) & raster & amplitude & full $\rightarrow0.25\times$ bus input & bus amplitude & $<.25$ \\
Overlap saturation (\odr) & routing & saturation & single $\rightarrow$ near-overlap route & near-overlap service & $[.35,.65]$ \\
Intensity saturation (\bis) & raster & amplitude & full $\rightarrow4\times$ bus input & bus amplitude & $[.35,.65]$ \\
\bottomrule
\end{tabularx}
\end{table}

Section~\ref{sec:reference_checks} and Appendix~\ref{app:routing_validation} explain the reference
checks and their applicability. Passing describes supported cases in this finite corpus, not
general correctness or optimizer safety.

\subsection{Architectures and Experimental Design}
\label{sec:arch}

The architecture contrast is ordered by how the decision variable is represented. \ufone\ and
\raone\ concatenate bus, rail, and opportunity inputs before one encoder. \dstwo\ gives the bus
intervention one encoder and concatenated rail/opportunity context a second. \tsthree\ further
separates rail from opportunities. Encoders compress spatial information, skip connections retain
local detail, and a shared decoder reconstructs the field. Shared residual components combine
several spatial scales; Appendix~\ref{app:architecture} gives their implementation. The experimental
axis is when the bus intervention and fixed context are fused.

At base width 128, the four families contain 31.9M, 42.4M, 71.2M, and 99.8M parameters. The
capacity-controlled comparison widens \ufone, \raone, and \dstwo\ to 224, 200, and 152 channels,
respectively, yielding approximately 100M parameters while \tsthree\ remains at width 128. Native
and widened pairs probe whether capacity alone explains an architectural difference.

The training comparison uses a five-step cumulative \tsthree\ sequence and an untreated/treated
\dstwo\ pair. The architecture comparison uses four families under common training, including
native-width and capacity-matched variants. Each controlled configuration has three independently
initialized models (Table~\ref{tab:design}). The weak-activation replication includes its three controlled-comparison runs
and seven other initializations (Appendix~\ref{app:seed_extension}); auxiliary ablations are
described in Appendix~\ref{app:results}.

The cumulative ladder adds geometric flips/rotations, mild continuous bus scaling, isolated-stop
examples with zero targets, and disconnected stub-edge examples. The scaling intervention targets
weak activation, isolated-stop examples target isolated-stop leakage, and stub-edge examples target
stub-edge leakage. These names describe training
interventions, not guaranteed outcomes. Configuration-level statements report the three seed
values and their mean $\pm$ standard deviation. The 50-set bootstrap describes uncertainty of one
checkpoint's corpus estimate and never increases the model-level sample size.

\section{Results}
\label{sec:experiments}

\subsection{Checking the Tests Against the Routing Calculation}
\label{sec:reference_checks}

To interpret a failed test as a surrogate weakness, we need evidence that the expected behavior
holds in the deterministic accessibility calculation it approximates. We therefore check the
test relations against that reference system, distinguishing three classes of evidence.

For \emph{routing-based checks}, we turn the generated scenes into routing jobs with corresponding
street networks, transit services and opportunity locations. The calculated responses pass the
overlap, spatial-localization and opportunity-displacement tests in every supported case across
the 50 sets. For \emph{analytical checks}, the isolated-stop, stub-edge and rail-context probes
have no effective bus service in their constructed networks. Their added accessibility is
therefore zero, satisfying the tests wherever the routed reference has sufficient signal;
these probe responses are derived analytically, not separately routed. The two remaining
relations are \emph{pixel-space stress tests}: scaling raster values does not specify a unique
change to transit service.

The first two classes ground the expected responses in the reference calculation, supporting
the interpretation of surrogate failures as departures from its behavior under the tested
conditions. Appendix~\ref{app:routing_validation} reports the results and the limits of the
correspondence between generated scenes and routing jobs.

\subsection{Structural Responses Remain Relation-Specific}

\begin{figure}[t]
  \centering
  \includegraphics[width=\textwidth]{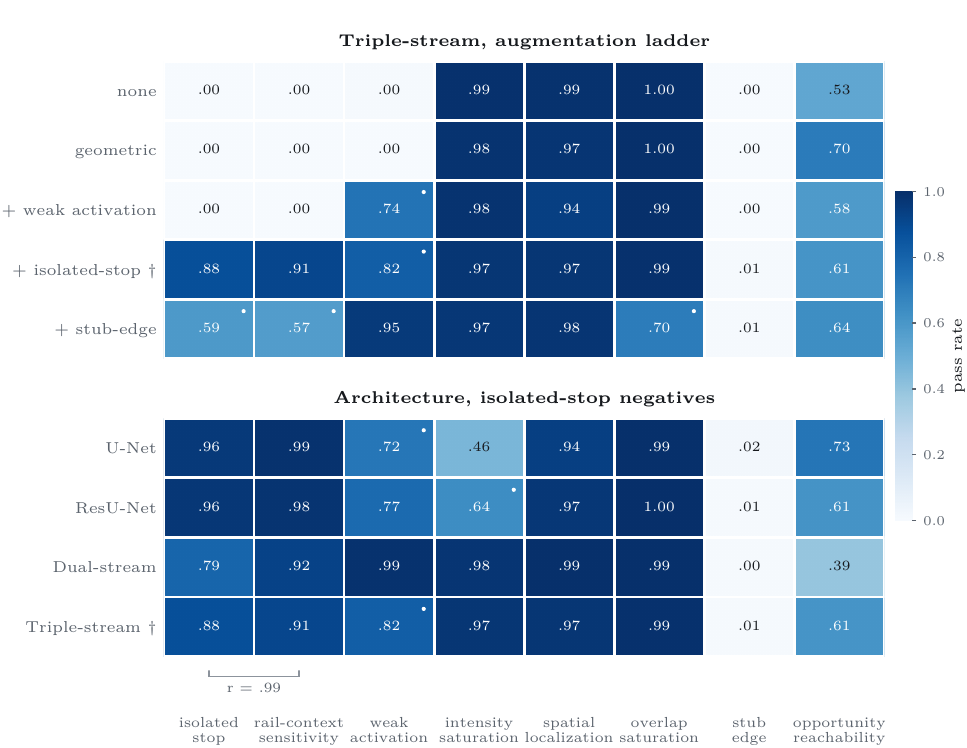}
  \caption{%
    \textbf{The generated suite distinguishes relation-specific responses.}
    Daggers mark a configuration repeated across blocks; dots mark seed ranges above 0.20.
  }
  \label{fig:regime_map}
\end{figure}

Figure~\ref{fig:regime_map} summarizes responses across the eight relations.
Connectivity and weak-activation responses vary with training data, intensity
saturation with architecture, and spatial localization remains high. Overlap saturation remains
high except after stub-edge training. Stub-edge pass rates remain low. Cells average three independently initialized models.

The surrogates do not match the routing calculation's opportunity response consistently:
configuration-mean pass rates range from $0.165$ to $0.730$. Only $0.295$--$0.564$ of eligible
cases have sufficient predicted reference signal to be scored, so pass rates alone cannot rank
the models. Appendix~\ref{app:opportunity_correction} reports this coverage, signed responses
and a stricter criterion that also requires a nonnegative remaining contribution.

The unaugmented \tsthree\ and \dstwo\ models have predictive skill $0.298\pm0.180$ and
$0.407\pm0.127$, but isolated-stop pass rates of only $0.003\pm0.001$ and $0.002\pm0.001$.
They beat the zero predictor while almost never satisfying the disconnected-stop relation.

Spatial-localization means remain uniformly high, $0.935$--$0.994$, and provide little
discrimination. Overlap-saturation means are similarly high, $0.991$--$0.997$, in every
configuration except the stub-edge rung, $0.702\pm0.198$.

Positive pooled skill does not imply good prediction in both regions: the unaugmented and
geometric triple-stream models have negative mean skill in Mecklenburg-Vorpommern. Isolated-stop
training improves mean prediction error in both regions. Separately, threshold sweeps preserve
the main connectivity and intensity comparisons, although absolute pass rates and weak activation
remain threshold-sensitive (Appendix~\ref{app:review_robustness}).

\subsection{Targeted Training Produces Specific and Uneven Responses}

\begin{figure}[t]
  \centering
  \includegraphics[width=\textwidth]{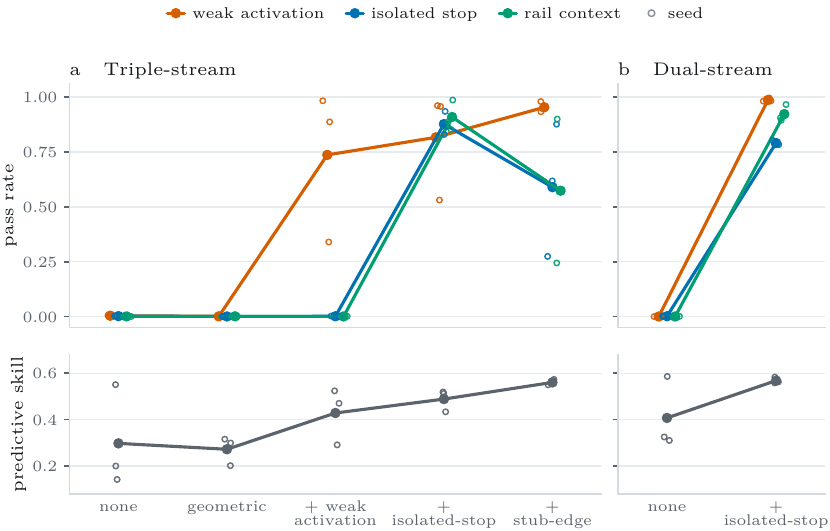}
  \caption{%
    \textbf{Training interventions produce specific and uneven responses.}
    Open circles show checkpoints and filled markers show their three-checkpoint means.
  }
  \label{fig:interventions}
\end{figure}

The cumulative ladder reveals three distinct outcomes (Figure~\ref{fig:interventions}).
Weak-activation training changes its intended relation: pass rate rises from
$0.005\pm0.002$ without augmentation to $0.737\pm0.283$. The three weak-activation checkpoints span
$0.340$--$0.983$, however, so the mean does not support a seed-stable improvement claim.
Across ten initializations, including these three, pass rates span
$0.091$--$0.985$ (Appendix~\ref{app:seed_extension}). Geometric
augmentation alone produces no material increase in weak-activation or connectivity pass rates.

Adding isolated-stop negatives produces the strongest result. In \tsthree, isolated-stop pass rate
rises from $0.002\pm0.001$ to $0.877\pm0.043$, rail-context sensitivity rises from $0.001\pm0.001$ to
$0.909\pm0.057$, and predictive skill rises from $0.428\pm0.099$ to $0.488\pm0.039$. All three
treated isolated-stop rates lie between $0.830$ and $0.935$.

The combined weak-activation and isolated-stop treatment produces a similar response in \dstwo:
isolated-stop and rail-context pass rates rise from $0.002\pm0.001$ and $0.001\pm0.000$ to
$0.791\pm0.009$ and $0.922\pm0.031$, respectively. Predictive skill rises from $0.407\pm0.127$ to
$0.568\pm0.011$. This extension shows that the connectivity improvement is not confined to one
stream layout, but it does not isolate the contribution of either treatment component.

Stub-edge negatives do not improve their target relation: its pass rate is $0.007\pm0.003$,
against $0.006\pm0.001$ at the isolated-stop rung. Isolated stop and rail context fall from $0.877$ and
$0.909$ to $0.590$ and $0.574$, and two of three seeds regress sharply. Overlap saturation falls
from $0.991$ to $0.702$ and fails in two of three seeds. Weak activation rises from $0.817$ to
$0.953$ and skill from $0.488$ to $0.560$. The intervention is ineffective on its target and
unstable in its side effects.

\subsection{Architecture Is Associated with Intensity Saturation, Not Overall Skill}

At matched capacity, intensity saturation cleanly separates stream groups. The mean pass rates are
$0.459\pm0.035$ for \ufone, $0.637\pm0.091$ for \raone, $0.975\pm0.011$ for \dstwo, and
$0.972\pm0.023$ for \tsthree. All multi-stream checkpoints lie between $0.940$ and $0.991$;
single-stream checkpoints lie between $0.409$ and $0.757$. Widening \ufone\ from 128 to 224
channels changes the mean from $0.383$ to $0.459$, and widening \raone\ from 128 to 200 changes it
from $0.394$ to $0.637$. Capacity narrows the gap but does not close it.

The overlap relation does not reproduce the architecture separation observed for intensity
saturation. Its means are $0.994$, $0.997$, $0.992$, and $0.991$ for the same four matched families.
Only intensity saturation supports an association between separating the bus intervention from
fixed context and reducing undesirable sensitivity to artificial input scaling. Because the
families also differ in encoder and fusion geometry, the comparison does not identify which
component causes the difference.

Figure~\ref{fig:skill_behavior} compares individual models, including the input-normalization
ablation described in Appendix~\ref{app:results}.
Across the comparable bus-induced-response checkpoints, multi-stream intensity-saturation rates
remain in one high band over nearly the full observed skill range, while single-stream rates form a disjoint lower
band. Isolated-stop behavior instead separates mainly by whether targeted negatives were present.
The x-axis alone identifies neither outcome.

\begin{figure}[t]
  \centering
  \includegraphics[width=\textwidth]{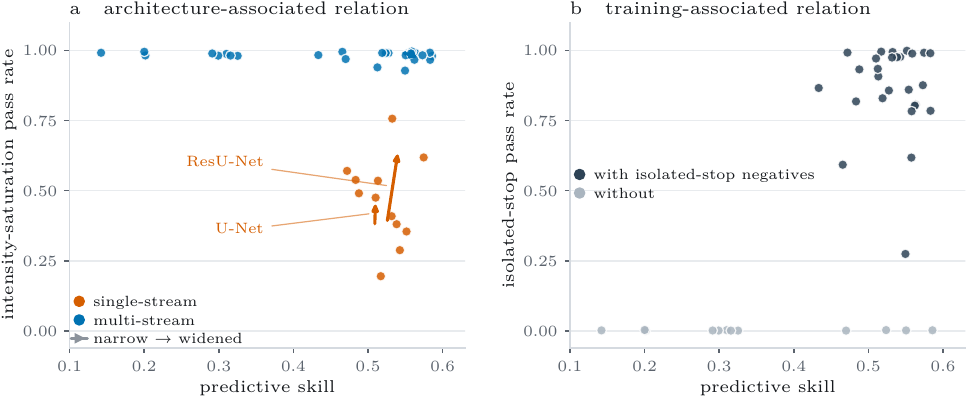}
  \caption{%
    \textbf{Predictive skill does not identify structural response.}
    Intensity saturation tracks architecture, whereas isolated-stop behavior tracks training data.
  }
  \label{fig:skill_behavior}
\end{figure}

\subsection{Aggregate Error Hides Large Category Differences}

The capacity-matched \dstwo\ treatment has the highest mean predictive skill among the controlled
configurations, $0.568\pm0.011$, with raw-space MAE $1.903\pm0.021$. It is the per-channel
showcase, not an overall model selection: its stub-edge response remains poor and its
opportunity response has incomplete support and substantial residual violations.

\begin{figure}[t]
  \centering
  \includegraphics[width=\textwidth]{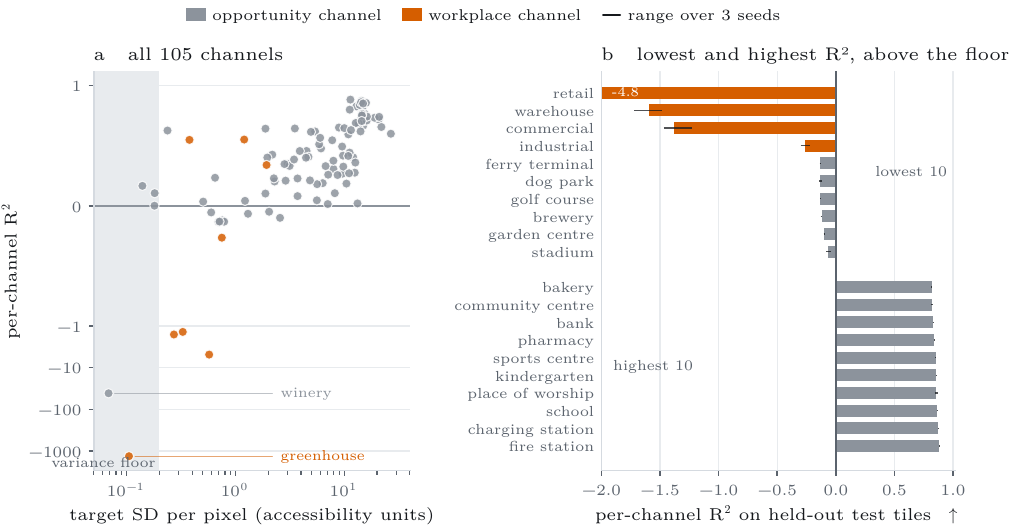}
  \caption{%
    \textbf{Predictive agreement varies across raw output channels.}
    Low-variance channels are shown but excluded from the named ranking.
  }
  \label{fig:channel_r2}
\end{figure}

Across its three checkpoints, median channel $R^2$ is $0.408$; 91 of 105 channels are above zero
and 44 exceed $0.5$. Fire-station accessibility is highest at $R^2=0.884$. Results are weaker for
workplace layers: median $R^2$ is $-0.823$ across the eight workplace channels, compared with
$0.418$ across 97 opportunity channels. This contrast is descriptive because category prevalence
and target variance differ substantially.
Figure~\ref{fig:channel_r2} shows all channels against target variation and ranks only the 100
above the standard-deviation floor of 0.2.

The extreme negative tail is amplified by low target variance. Winery target SD is $0.068$ and mean
$R^2=-41.2$; greenhouse target SD is $0.105$ and mean $R^2=-1338.8$. Across all channels,
target SD and $R^2$ have Spearman correlation $+0.76$. The variance floor keeps the five affected
channels visible without allowing this metric pathology to define the ranking.

\subsection{Per-Candidate Computational Cost}
With fixed baselines reused, four candidates across two development tiles required median
reference times of 113--129\,s, versus 40--54\,ms for complete GPU surrogate evaluation and
612--632\,ms on CPU. Excluding candidate preparation and unprofiled process overhead still
gives reference-to-GPU ratios of approximately 1,850--2,360. Appendix~\ref{app:runtime} reports
hardware, repetitions and timing boundaries. This implementation benchmark supports the
computational motivation, not downstream optimization benefit.

\FloatBarrier
\section{Discussion}
\label{sec:discussion}

Predictive agreement and structural response provide different evidence about a surrogate.
Models can beat the zero predictor while responding incorrectly to disconnected service,
consistent with scientific-model studies of implausible responses despite conventional accuracy
\citep{reichert2024metamorphic}. Targeted isolated-stop examples improve both connectivity and
prediction, illustrating the value of failure-directed training
\citep{dreossi2018counterexample}. The combined dual-stream treatment supports transfer across
families but does not isolate its individual training components.

Neither training nor architecture provides a general repair. Weak activation remains unstable
across ten initializations; stub-edge training fails; and opportunity displacement retains support
and signed-response limitations. Multi-stream models perform better on intensity saturation,
but not overlap saturation. Similar dual- and triple-stream responses do not establish a need
for separate rail and opportunity encoders, and other architectural differences prevent causal
attribution to stream separation.

The tests also have limits. Generated sets can share artifacts, and calibration draws on the
same national data source as training. Thresholds were fixed before evaluation but not externally
preregistered. Reference networks support the tested premises without reproducing every procedural
tensor exactly; analytical controls and raster-amplitude probes provide different evidence from
routed responses. Conditional pass rates require coverage information, and set-cluster intervals
describe corpus variation rather than independent model replication.

The inputs are real geographic data, while the targets are counterfactual values from the reference
calculation. Predictive evaluation therefore measures agreement with that calculation, not with
observed accessibility outcomes. The 818 test tiles belong to only two regions and remain
spatially dependent, limiting claims about geographic transfer
\citep{roberts2017crossvalidation,ploton2020spatial}. Low target variance also limits interpretation
of per-channel $R^2$.

Finally, behavioral repair and lower latency do not establish better optimizer-selected networks.
That comparison requires fixed admissible designs, search, objective weights and budget:
connected-route constraints can exclude tested failures, while unrestricted rasters may lack
routing counterparts. Passing a finite suite is not a guarantee of optimizer safety.

\section{Conclusion}
\label{sec:conclusion}

Accessibility surrogates should be evaluated through both geographic prediction error and
controlled behavioral tests. Targeted training can reduce specific violations without sacrificing
prediction, but improvements do not extend to every relation. Checking test premises against
the reference calculation makes these behavioral findings more interpretable.

\paragraph{Code and data availability.}
The companion code repository (\url{https://github.com/tillzacher/transit-accessibility-surrogate}) provides the scenario generator,
frozen definitions and calibration, scoring code, recorded training configurations, model and
training code, tile-construction code, and per-model results for score-level reanalysis. It also
provides synthetic routing inputs and reference extraction code. The accessibility pipeline that
produces the reference calculations is released separately as MAPPER v2.0.0
(\url{https://github.com/TUMFTM/MAPPER}, \url{https://doi.org/10.5281/zenodo.23557362}).
Trained checkpoints are archived at Zenodo (\url{https://doi.org/10.5281/zenodo.22750422}).
Original regional training tensors and the source database are not redistributed; they can be
rebuilt from open data with the released pipelines. The repository documents these reproduction
boundaries explicitly.

\subsubsection*{Acknowledgments}
This research was funded by the Federal Ministry of Research, Technology and Space (BMFTR) as part
of the MCube Cluster (Clusters4Future) within the project DatSim2.0 (grant no.\ 03ZU2105HA).
The authors gratefully acknowledge the Leibniz Supercomputing Centre (LRZ) for providing compute
resources on the LRZ AI Systems, and the Hightech Agenda Bayern for supporting this infrastructure.
The funders had no role in study design, analysis, interpretation, or the decision to publish.

\bibliography{p2}
\bibliographystyle{tmlr}

\FloatBarrier
\appendix
\clearpage
\section{Architecture Detail}
\label{app:architecture}

The selected width-152 dual-stream model assigns the candidate bus intervention to one encoder
and concatenates rail and opportunity tensors in a second, fixed-context encoder. Each encoder
progressively increases its spatial receptive field while skip connections preserve local detail
for the decoder. Gated $1\times1$ fusions control how much information crosses between streams at
each resolution. The streams fully merge at a bottleneck that combines atrous spatial pyramid
pooling (ASPP), which aggregates several spatial scales, with squeeze-and-excitation (SE), which
reweights feature channels. A shared decoder then emits 32 normalized PCA channels. The
triple-stream family separates rail and opportunity contexts, whereas the single-stream families
concatenate all inputs before one encoder.

\begin{figure}[h]
  \centering
  \includegraphics[width=\textwidth]{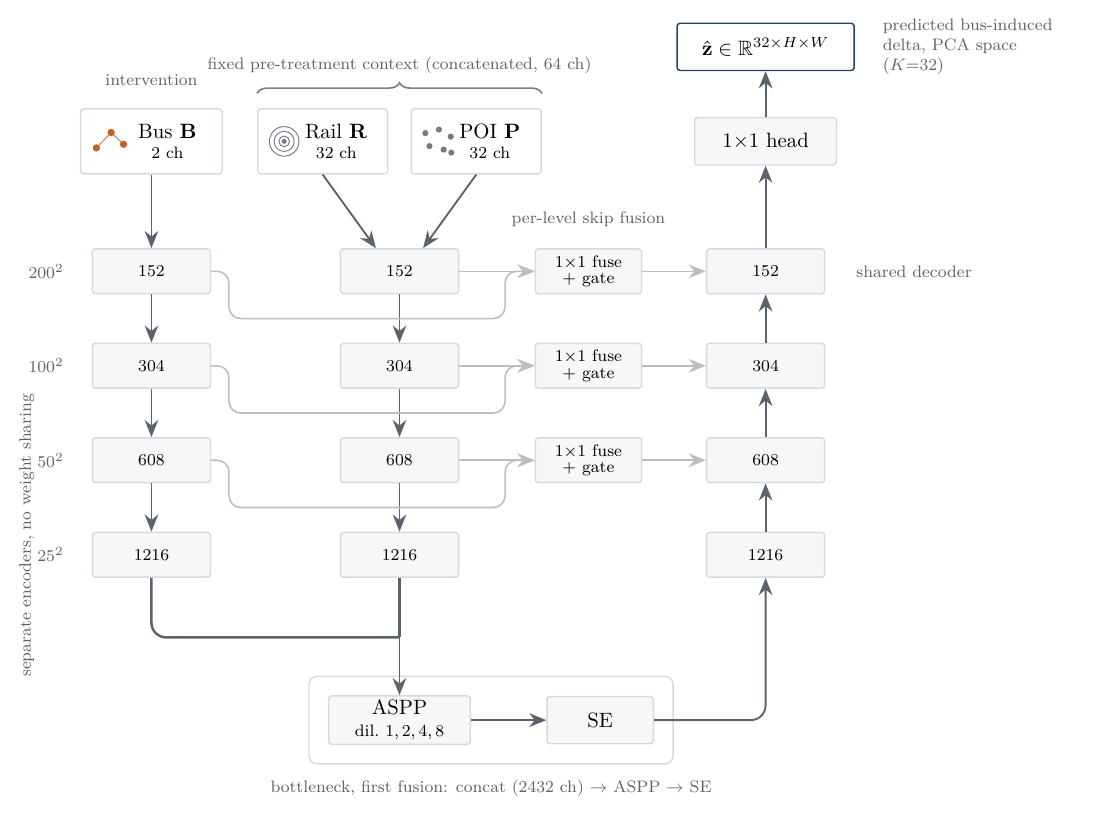}
  \caption{%
    \textbf{Dual-stream architecture used for the per-channel analysis.}
    The bus intervention remains separate from fixed context until feature fusion.
  }
  \label{fig:architecture}
\end{figure}
\FloatBarrier

\section{Reference-Calculation Validation}
\label{app:routing_validation}

\paragraph{Constructed network counterparts.}
For each of the 50 procedural sets, we construct street graphs, timetables and opportunity
sources from the recorded scene geometry. Opportunity conditions $A$, $B$ and $N$ retain the
same streets, rail, bus service and evaluation mask. Only the focal opportunity source changes.
Sources include amenity points and workplace footprints; total employment is a derived aggregate
and is excluded from the independently moved-focal comparison. Known-empty background sources
are distinguished from missing output columns.

The calculation uses an exponential travel-time kernel with parameter $0.0005\,\mathrm{s}^{-1}$,
a 3,500-s cutoff and ceiling 100. H3-resolution-10 fields are sampled at pixel centres on a
$200\times200$ grid with 105-m cells in EPSG:32632. This is 105 ground metres, whereas the training
grid uses projected metres in EPSG:3857. Service conventions, network boundaries and kernel
parameters are fixed within each comparison. Separate condition caches prevent reuse of another
condition's graph. The artifact records the input, configuration and executable fingerprints.

The correspondence audit covers all 300 geometry-family/set combinations. Reconstructed
opportunity and stop rasters differ by at most $1.14\times10^{-6}$; bus-edge support intersection
over union is at least $0.855$, with centreline displacement at most one pixel.
Bus-edge amplitude is not matched, and rail context is calculated rather than the original
synthetic field. These are explicit scene counterparts, not exact full-input replays.
Surrogate tests use the procedural tensors and their background-only controls.

\begin{table}[htbp]
\centering
\scriptsize
\setlength{\tabcolsep}{4pt}
\caption{\textbf{Reference checks across 50 synthetic sets. Intervals resample sets; channels remain nested within sets.}}
\label{tab:routing_validation}
\begin{tabular}{@{}llrrl@{}}
\toprule
Relation & Check & Pass/support & Pooled rate & 95\% interval \\
\midrule
Isolated stop (\isl) & Analytic zero probe & 3919/3919 & 1.000 & [1.000, 1.000] \\
Stub edge (\sel) & Analytic zero probe & 3854/3854 & 1.000 & [1.000, 1.000] \\
Rail context (\rcs) & Analytic zero probe & 2853/2853 & 1.000 & [1.000, 1.000] \\
Overlap (\odr) & Calculated & 3927/3927 & 1.000 & [1.000, 1.000] \\
Localization (\sdl) & Calculated & 3913/3913 & 1.000 & [1.000, 1.000] \\
Interchange-only diagnostic & Calculated & 551/2853 & 0.193 & [0.155, 0.229] \\
Opportunity (\prs) & Calculated & 3998/3998 & 1.000 & [1.000, 1.000] \\
\bottomrule
\end{tabular}
\end{table}

For opportunity attenuation, 3,998 of 4,024 eligible effects reach the support floor,
and all supported effects pass. The largest supported absolute residual ratio is $0.0507$.
There are nevertheless 263 supported negative probe increments, with the most negative raw
increment $-5.39\times10^{-5}$; requiring an exactly nonnegative probe increment reduces the
joint pass rate to $0.934$. We retain these signs rather than declaring exact monotonicity from
attenuation. The corpus bootstrap is degenerate at one when every supported observation passes;
it is not a bound on unseen cases.

Isolated-stop, endpoint-stub and bus-edge-removal probes contain no effective bus service in the
constructed counterpart. Their bus-induced response is analytically zero, not a separately
measured routing success. References are routed and must meet the support floor. This control
does not establish that arbitrary short or dead-end services have no value. A separate diagnostic
that removes only a rail interchange is not the implemented rail-context test.
Its pass rate is $0.193$, illustrating why this different intervention cannot inherit the
bus-edge-removal premise. Likewise, scalar
raster multiplication does not specify a unique timetable or route change; the two amplitude
probes remain input-space diagnostics.

These checks concern the constructed procedural scenes. Passing their operational gates does
not establish universal relations across real networks, including for spatial localization.
The validation does not evaluate optimizer-selected network quality.
\FloatBarrier

\section{Opportunity-Measurement Correction}
\label{app:opportunity_correction}

The original corpus contains 13 distinct scenes per set, with shared references supporting eight
relations. These 650 scenes were frozen before comparing the reported models. The retrospective
measurement correction adds one background-only control per set, yielding 14 scenes per set
without changing the original reference or modified tensors.

The original two-condition score and all original predictions are retained in the analysis
archive. Reference validation identified background dilution in that score: with unchanged
background gain $G$ and focal gain $F$, perfect removal of the focal contribution gives
$(G+F)/(2G+F)$, which fails the $0.75$ gate unless $F>2G$. The original near-zero pass rates
therefore do not support an interpretation of opportunity blindness.
The revised calculation evaluates the original near/displaced tensors together with a
background-only control. It changes neither trained weights nor geographic prediction results.
Negative increments retain their magnitude penalty and a separate strict-sign diagnostic; they
are never converted to perfect attenuation by clipping.

\begin{table}[htbp]
\centering
\scriptsize
\setlength{\tabcolsep}{4pt}
\caption{\textbf{Corrected opportunity response and support on the full raster schema. Means over three models per row.}}
\label{tab:opportunity_correction}
\begin{tabular}{@{}lrrrrr@{}}
\toprule
Configuration & Coverage & Pass/support & Pass/eligible & Strict joint & Negative probe \\
\midrule
Triple, none & 0.475 & 0.528 & 0.255 & 0.197 & 0.616 \\
Triple, geometric & 0.518 & 0.705 & 0.371 & 0.358 & 0.488 \\
Triple, + weak activation & 0.489 & 0.583 & 0.288 & 0.303 & 0.416 \\
Triple, + isolated stop & 0.564 & 0.609 & 0.351 & 0.446 & 0.267 \\
Triple, + stub edge & 0.563 & 0.636 & 0.362 & 0.412 & 0.360 \\
Dual, none & 0.314 & 0.165 & 0.047 & 0.106 & 0.109 \\
Dual, treated (matched) & 0.494 & 0.392 & 0.193 & 0.267 & 0.193 \\
Dual, treated (native) & 0.452 & 0.416 & 0.187 & 0.276 & 0.217 \\
U-Net, treated (native) & 0.295 & 0.582 & 0.178 & 0.356 & 0.343 \\
U-Net, treated (matched) & 0.474 & 0.730 & 0.347 & 0.436 & 0.398 \\
ResU-Net, treated (native) & 0.454 & 0.527 & 0.242 & 0.255 & 0.429 \\
ResU-Net, treated (matched) & 0.557 & 0.610 & 0.341 & 0.341 & 0.431 \\
Triple, raw-bus ablation & 0.685 & 0.785 & 0.535 & 0.458 & 0.403 \\
\bottomrule
\end{tabular}
\end{table}

Coverage is the fraction of generated focal channel/set observations whose positive reference
increment reaches the unchanged floor. Conditional pass rate uses only that supported subset;
the all-eligible passing fraction keeps the size of that subset visible. A zero-output model has
zero coverage, undefined conditional pass rate and zero all-eligible passing fraction. Negative
probe fractions and strict joint rates are conditional on support. Their differences show why
attenuation should not be equated with monotonicity. The table uses the full 105-channel raster
schema. An additional reference-grounded comparison in the artifact excludes the independently
unrepresentable total-employment focal and uses the same eligible subset for all models.
This restriction changes individual-model conditional pass rates by at most 0.0028.

\section{Complete Experimental Results}
\label{app:results}

Tables~\ref{tab:ladder_results} and \ref{tab:architecture_results} report the numbers underlying
Figure~\ref{fig:regime_map}. Every entry is a mean over three independently initialized model
instances. Pass rates pool supported set--channel observations within each model instance before
the three instances are averaged. Uncertainty and unsupported observations follow the procedure in
Section~\ref{sec:eval_framework}.
For compactness, the tables use ISL for isolated stop, RCS for rail-context sensitivity, WAL for weak
activation, BIS for intensity saturation, SDL for spatial localization, ODR for overlap
saturation, SEL for stub edge, and PRS for opportunity reachability.

\begin{table}[h]
\centering
\scriptsize
\setlength{\tabcolsep}{2.4pt}
\caption{\textbf{Cumulative training ladder and dual-stream transfer.} Skill uses all 818
geographic test tiles. The remaining columns are supported-observation pass rates.}
\label{tab:ladder_results}
\begin{tabular}{@{}llccccccccc@{}}
\toprule
Family & Training & Skill & \isl & \rcs & \wal & \bis & \sdl & \odr & \sel & \prs \\
\midrule
\tsthree & none & .298 & .003 & .002 & .005 & .990 & .987 & .995 & .002 & .528 \\
& geometric & .272 & .001 & .002 & .002 & .982 & .974 & .996 & .002 & .705 \\
& + weak activation & .428 & .002 & .001 & .737 & .983 & .935 & .994 & .003 & .583 \\
& + isolated-stop & .488 & .877 & .909 & .817 & .972 & .971 & .991 & .006 & .609 \\
& + stub-edge & .560 & .590 & .574 & .953 & .967 & .976 & .702 & .007 & .636 \\
\midrule
\dstwo & none & .407 & .002 & .001 & .001 & .983 & .985 & .997 & .001 & .165 \\
& + weak activation, isolated-stop & .568 & .791 & .922 & .987 & .975 & .994 & .992 & .005 & .392 \\
\bottomrule
\end{tabular}
\end{table}

\begin{table}[h]
\centering
\scriptsize
\setlength{\tabcolsep}{2.4pt}
\caption{\textbf{Architecture and capacity results under the common isolated-stop treatment.}
The matched rows are approximately 100M parameters. The \tsthree\ matched row uses the same
three checkpoints as the isolated-stop rung in Table~\ref{tab:ladder_results}.}
\label{tab:architecture_results}
\begin{tabular}{@{}llccccccccc@{}}
\toprule
Family & Width & Skill & \isl & \rcs & \wal & \bis & \sdl & \odr & \sel & \prs \\
\midrule
\ufone & 128 & .509 & .988 & .978 & .790 & .383 & .948 & .997 & .003 & .582 \\
\ufone & 224 matched & .510 & .960 & .987 & .723 & .459 & .935 & .994 & .018 & .730 \\
\raone & 128 & .526 & .932 & .985 & .660 & .394 & .968 & .997 & .046 & .527 \\
\raone & 200 matched & .540 & .965 & .980 & .772 & .637 & .965 & .997 & .006 & .610 \\
\dstwo & 128 & .548 & .843 & .915 & .946 & .991 & .992 & .994 & .004 & .416 \\
\dstwo & 152 matched & .568 & .791 & .922 & .987 & .975 & .994 & .992 & .005 & .392 \\
\tsthree & 128 matched & .488 & .877 & .909 & .817 & .972 & .971 & .991 & .006 & .609 \\
\bottomrule
\end{tabular}
\end{table}

A separate bus-induced-response ablation leaves bus channels in their raw $[0,1]$ scale. Its mean
skill is $0.536$; its \isl, \rcs, \wal, and \bis\ pass rates are $0.857$, $0.938$, $0.949$,
and $0.995$. It is an input-normalization ablation, not a substitute for the common treated
cell. Another checkpoint predicts total accessibility and has skill $-0.259$ after conversion
to a bus-induced response. Because its target changes the meaning of the structural comparisons,
it is excluded from all probe contrasts and correlation analyses.

\begin{figure}[h]
  \centering
  \includegraphics[width=\textwidth]{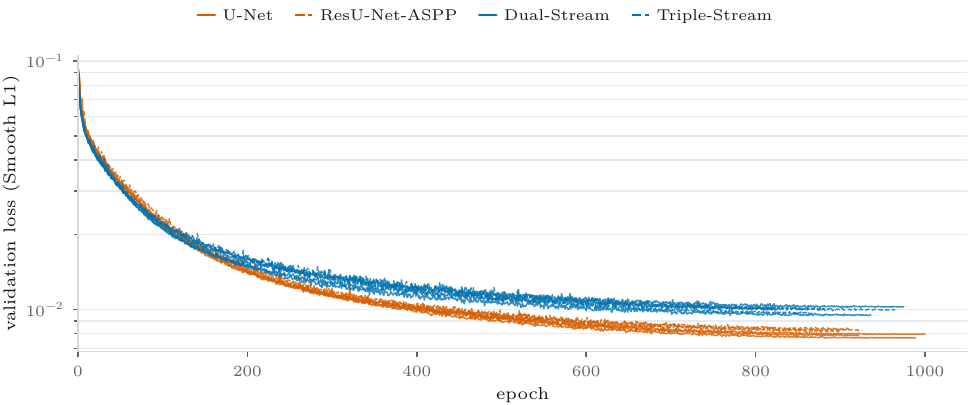}
  \caption{\textbf{Training convergence.} Validation-loss histories for three checkpoints in each
  capacity-matched family; checkpoint selection uses the best validation epoch.}
  \label{fig:convergence}
\end{figure}

\section{Weak-Activation Replication}
\label{app:seed_extension}

The weak-activation analysis uses ten initializations, including the three in the controlled
comparison. All share the same geographic split, validation membership, PCA and normalization.
Seeds vary weight initialization, minibatch ordering and stochastic augmentation, not the split.
Other controlled configurations use three initializations.

All ten models were evaluated on the same 50 procedural sets. The weak-activation measurement
uses a cutoff of 0.25. The combined conditional pass rate has mean 0.743,
sample standard deviation 0.351, median 0.944 and range 0.091--0.985. Restricting evaluation to
the 4,952 set--channel observations supported by every model yields mean 0.742 and range
0.084--0.986. Different support membership therefore does not explain the spread. This confirms
instability across training realizations; it does not identify which stochastic component
causes it or establish an inevitable failure of the architecture.

\begin{figure}[htbp]
\centering
\includegraphics[width=\textwidth]{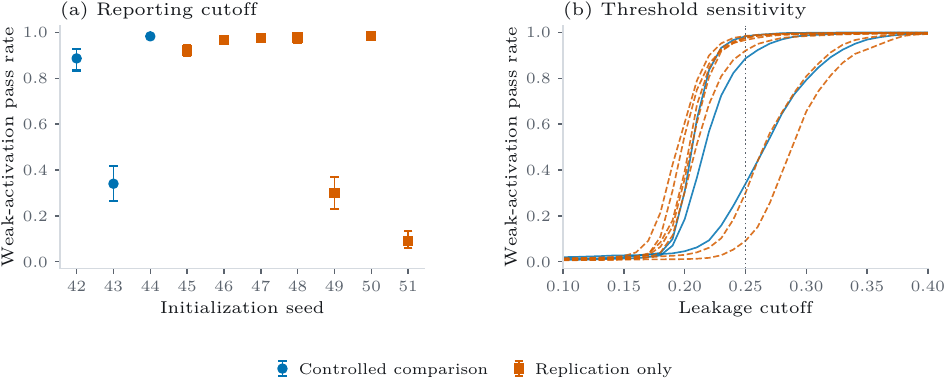}
\caption{Weak-activation replication across ten initializations. Left: individual pass rates
with 95\% set-cluster bootstrap intervals. Right: fixed-support threshold sweeps, with the
reporting cutoff dotted. Color and line style distinguish controlled-comparison and replication-only runs.}
\label{fig:seed_extension}
\end{figure}
\FloatBarrier

\section{Threshold Sensitivity and Regional Prediction}
\label{app:review_robustness}

We sweep leakage cutoffs from 0.10 to 0.40, sensitivity cutoffs from 0.60 to 0.90, and
saturation-window half-widths from 0.05 to 0.25 in increments of 0.01. Predictions and support
membership remain fixed within each sweep. Opportunity displacement uses background-controlled
attenuation. This analysis is descriptive, not threshold selection; the support floor is not swept.

The mean connectivity advantage of isolated-stop training persists throughout the tested
cutoffs: its isolated-stop advantage over the preceding rung ranges from 35.3 to 95.7 percentage
points, and its rail-context advantage from 33.0 to 97.3 points. Stub-edge training does not
repair its target: up to a cutoff of 0.30, its stub-edge pass rate exceeds that of the
isolated-stop rung by at most one percentage point. It lowers isolated-stop pass rates at every
cutoff and rail-context pass rates at every cutoff except the strictest, 0.90, where they rise by
1.2 points. It lowers overlap saturation by 29 to 37 points in every window. Both multi-stream
configurations outperform both matched single-stream configurations on intensity saturation
throughout the tested windows, with a minimum mean margin of 6.35 percentage points. Absolute
pass fractions nevertheless change substantially, and weak activation remains both seed- and
threshold-sensitive. These comparisons concern configuration means, not seed-by-seed dominance.

\begin{table}[htbp]
\centering
\scriptsize
\setlength{\tabcolsep}{4pt}
\caption{\textbf{Regional prediction. Means over three initializations; Munich has 81 tiles and Mecklenburg-Vorpommern 737.}}
\label{tab:regional_prediction}
\begin{tabular}{@{}lrrrr@{}}
\toprule
Configuration & Munich MAE & Munich skill & MV MAE & MV skill \\
\midrule
Triple, none & 9.816 & 0.790 & 1.611 & -0.059 \\
Triple, geometric & 9.710 & 0.803 & 1.677 & -0.113 \\
Triple, + weak activation & 9.574 & 0.819 & 1.358 & 0.145 \\
Triple, + isolated stop & 9.526 & 0.825 & 1.235 & 0.244 \\
Triple, + stub edge & 9.639 & 0.812 & 1.069 & 0.378 \\
Dual, none & 9.690 & 0.805 & 1.391 & 0.118 \\
Dual, treated (matched) & 9.607 & 0.815 & 1.056 & 0.388 \\
Dual, treated (native) & 9.596 & 0.817 & 1.099 & 0.354 \\
U-Net, treated (native) & 9.627 & 0.813 & 1.179 & 0.289 \\
U-Net, treated (matched) & 9.695 & 0.805 & 1.170 & 0.296 \\
ResU-Net, treated (native) & 9.496 & 0.829 & 1.158 & 0.306 \\
ResU-Net, treated (matched) & 9.536 & 0.824 & 1.123 & 0.334 \\
Triple, raw-bus ablation & 9.601 & 0.816 & 1.125 & 0.333 \\
\bottomrule
\end{tabular}
\end{table}

\FloatBarrier

Regional skill uses the regional zero-prediction and PCA-reconstruction errors in
Equation~\ref{eq:skill}. These reference MAEs are 16.278 and 8.097 in Munich, versus 1.538 and
0.297 in Mecklenburg-Vorpommern. Tile-count-weighted regional MAEs reproduce pooled MAEs.
The unaugmented and geometric triple-stream configurations have negative mean skill in the
rural holdout despite positive pooled skill. Isolated-stop training improves mean MAE in both
regions. Adding stub-edge training then worsens it in Munich, from 9.526 to 9.639, but improves
it in Mecklenburg-Vorpommern, from 1.235 to 1.069. The matched dual-stream configuration
has the lowest rural mean MAE, and the native-width residual configuration the lowest Munich mean MAE.
There is no universal regional winner. Training-seed variation does not estimate uncertainty
over new regions, and the procedural behavioral corpus has no Munich/rural membership.
\FloatBarrier

\begin{figure}[p]
\centering
\includegraphics[width=\textwidth]{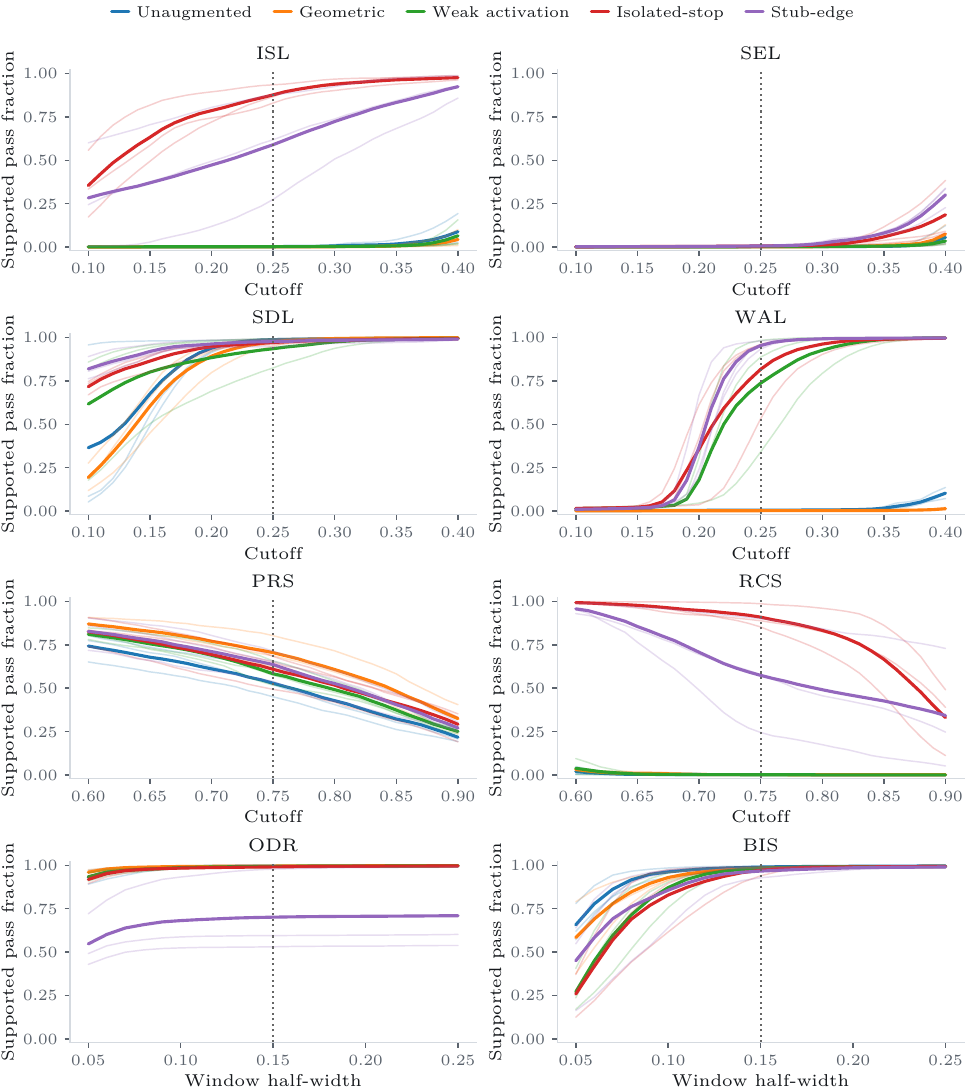}
\caption{\textbf{Threshold sensitivity of the training ladder.} Thin lines show individual
initializations and thick lines configuration means; dotted lines mark the reporting cutoffs.
Opportunity reachability uses background-controlled attenuation.}
\label{fig:threshold_ladder}
\end{figure}
\begin{figure}[p]
\centering
\includegraphics[width=\textwidth]{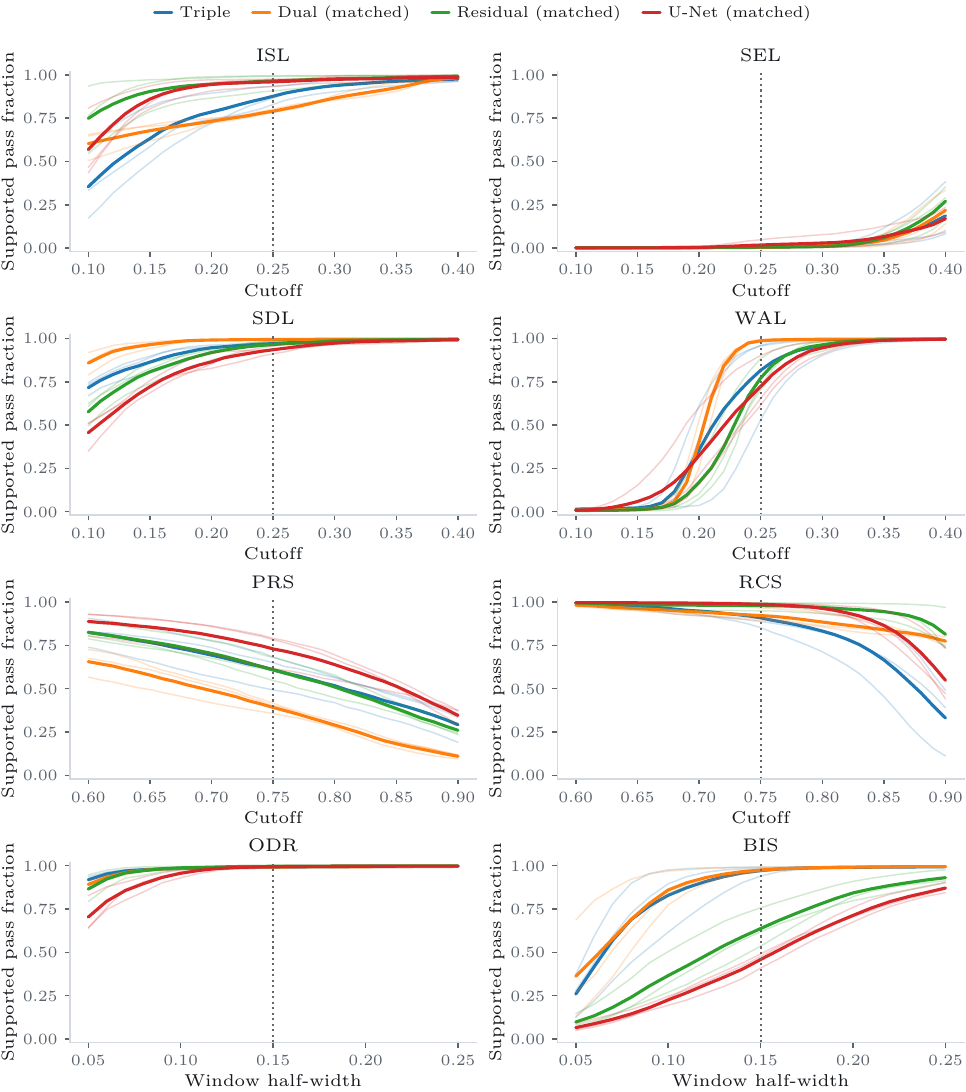}
\caption{\textbf{Threshold sensitivity of the capacity-matched architectures.} Individual
initializations and configuration means use the same convention as Figure~\ref{fig:threshold_ladder}.}
\label{fig:threshold_architectures}
\end{figure}
\FloatBarrier

\section{Matched Per-Candidate Runtime}
\label{app:runtime}

We benchmarked one \tsthree\ model after isolated-stop training on two
Schleswig-Holstein development tiles, each with a single-line and overlapping-line candidate.
These four cases are not a representative geographic sample and are
separate from both the geographic prediction holdouts and the procedural test corpus.
Both implementations return a host-resident, 105-channel $200\times200$ accessibility-gain
field. The reference retains the surrounding 80-km buffered network rather than routing
only inside the output tile. Fixed rail and opportunity rasters use the same
source snapshot as the reference calculation.

The fixed walking/rail baseline, street graph and raw opportunity extraction are reused.
Reference timing starts with the candidate route description and includes timetable generation
and merging, transit-graph construction, road/transit connection, source snapping and workplace
extraction, fresh routing-index construction, accessibility accumulation, tile extraction and
baseline subtraction. Each repetition uses a private candidate graph and index; no final
candidate result is cached. Surrogate timing includes route rasterization, normalization,
device transfers, inference, inverse PCA and return of the raw field. Training, model loading
and shared context preparation are excluded. This measures the existing implementations,
not an incrementally updated routing algorithm or full cold-start data import.

The reference uses native C++ with RoutingKit customizable contraction hierarchies.
Both workloads ran sequentially on the same Intel Core i9-14900K host with 188.5\,GiB system
memory, sixteen requested CPU threads, and an NVIDIA RTX 4080 with 16\,GB memory.
Surrogate inference used PyTorch 2.7.1 and CUDA 11.8. We performed three routing repetitions
per candidate, shuffling candidate order within repetitions, and twenty batch-one surrogate
evaluations per device after ten warmups. GPU timing synchronizes device completion.
Repeated baseline and candidate reference calculations produce identical outputs. Technical repetitions quantify timing variability, not independent geographic evidence.

\begin{table}[htbp]
\centering
\small
\setlength{\tabcolsep}{5pt}
\caption{\textbf{Matched candidate latency.} Medians; reference brackets give the range of three runs.}
\label{tab:runtime}
\begin{tabular}{@{}llrrr@{}}
\toprule
Tile & Candidate & Reference, s & GPU, ms & CPU, ms \\
\midrule
1 & Single line & 128.88 [127.55--129.44] & 42.45 & 611.73 \\
1 & Overlapping lines & 128.70 [128.61--129.56] & 53.87 & 631.88 \\
2 & Single line & 113.13 [112.80--113.72] & 39.66 & 613.74 \\
2 & Overlapping lines & 113.44 [113.13--116.97] & 48.24 & 627.87 \\
\bottomrule
\end{tabular}
\end{table}

Full per-candidate reference-to-surrogate ratios of medians range from 2,352 to 3,036 on GPU
and 181 to 211 on CPU. Native process wall time includes idle progress-thread shutdown.
A secondary boundary uses the engine's completed-work timer, ending after output-file
completion, plus tile extraction.
This conservative secondary boundary excludes candidate feed/graph preparation and unprofiled
process overhead, while retaining complete surrogate evaluation. Its ratios are 1,849--2,358
on GPU and 142--164 on CPU; the native timer has 0.01-s resolution.
One-time baseline process times were 90.06 and 105.05\,s, with extraction taking 1.84 and
0.54\,s and context construction a further 3.08 and 3.36\,s. These preparation observations
are not a full cold-start benchmark. All ratios are specific to the stated hardware, model,
network scope and cache policy; they establish neither a universal routing speedup nor improved
optimizer-selected accessibility.
\FloatBarrier

\section{Testing-Instrument Reproducibility}
\label{app:instrument}

Generation statistics come from 1,000 training tiles and 100 network specifications.
The 50-set corpus balances variation, storage and evaluation cost rather than following a formal
power analysis. Six geometry families support the eight relations.
Each generated set uses independent random draws for isolated-stop leakage, opportunity
reachability, rail-context sensitivity, amplitude response, spatial localization, and stub-edge leakage.
WAL and BIS share the same amplitude reference tile but are evaluated with separate $0.25\times$
and $4\times$ input transformations.

\begin{table}[h]
\centering
\footnotesize
\caption{\textbf{Fourteen scenes per generated set.} WAL and BIS apply their scaling probes
to the same full-amplitude reference and therefore require only one base scenario.}
\label{tab:scenario_files}
\begin{tabularx}{\textwidth}{@{}lXl@{}}
\toprule
Family & Scenario pair or base & Invariant control \\
\midrule
\isl & isolated stop without opportunities; isolated and connected stops with opportunities
& bus edges removed; stops and context fixed \\
\prs & on-route, displaced and background-only opportunities & only focal POI changes \\
\rcs & stop near rail, with and without bus edges & stops and context fixed \\
\wal/\bis & connected route at full amplitude & only bus amplitude changes during evaluation \\
\odr & single route; near-overlapping route & only near-overlap edge mass differs \\
\sdl & localized route with near and far evaluation zones & equal-area, non-overlapping masks \\
\sel & origin stub; full route & only route extent differs \\
\bottomrule
\end{tabularx}
\end{table}

We verify tensor shape, channel alignment, border margins, mask equality, and numerical equality
of every nuisance tensor in named pairs. The rasterizer uses stop, edge, and POI
Gaussian widths of $1.0$, $0.65$, and $1.0$ pixels, respectively, followed by the same soft cap as
the supervised-data rasterization procedure. An independent rasterization check found maximum absolute disagreement of about
$6\times10^{-8}$. One of every five sets activates all 105 categories; the other four sample
empirical density and category frequencies estimated from training data. This schedule preserves
coverage of rare categories
without making every tile an unrealistic all-channel stress case.

\begin{figure}[htbp]
  \centering
  \includegraphics[width=0.70\textwidth]{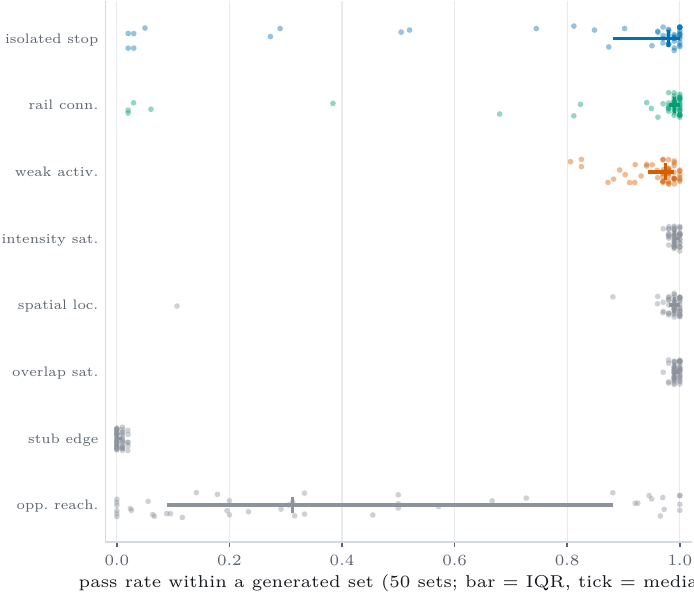}\\[4pt]
  \includegraphics[width=0.60\textwidth]{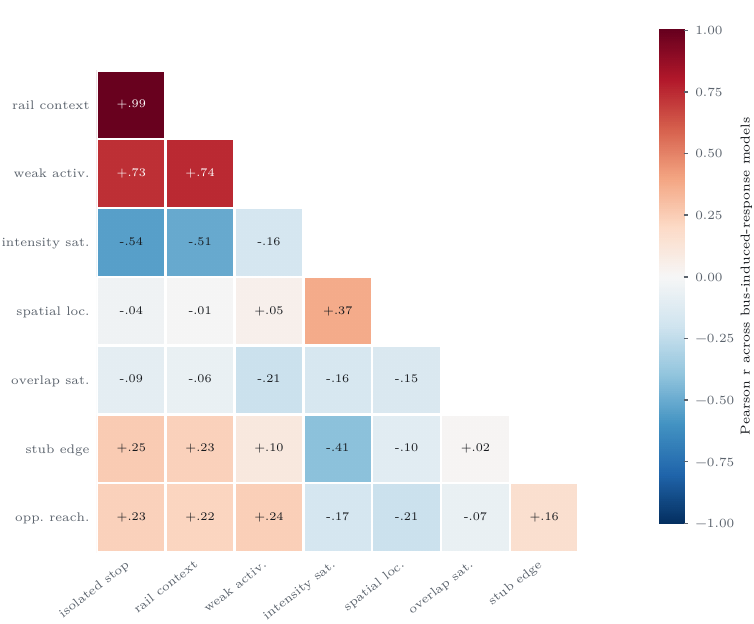}
  \caption{\textbf{Variation within the testing instrument.} Top, pass-rate variation over
  50 generated sets. Bottom, descriptive probe correlations over comparable checkpoints.}
  \label{fig:instrument_variation}
\end{figure}

In the upper panel of Figure~\ref{fig:instrument_variation}, each point is a within-set pass rate
for one treated \tsthree\ checkpoint; bars span the interquartile range and ticks mark medians.
These 50 sets measure corpus variation, not model replication. The lower panel compares relations
over the comparable checkpoints. Its correlations are descriptive because the experiment is structured
rather than sampled from a population of models; the strong \isl--\rcs\ association is consistent
with a shared connectivity interpretation without making the relations interchangeable.
Both probes remove bus edges while retaining stop and context rasters: isolated stop emphasizes
destination opportunities, whereas rail-context sensitivity emphasizes a rail-accessibility
peak. They therefore test one connectivity requirement in different contexts, not two independent
failure mechanisms. Their pass-rate correlation is $0.994$ in the 33-model controlled
comparison and $0.778$ within its common-treatment architecture subset. The matched dual-stream
model's distinct mean rates, $0.791$ and $0.922$, also show that one cannot substitute numerically
for the other. The figure uses pooled rates across comparable bus-induced-response models.

A ninth stop-density diagnostic was excluded before the reported analysis. In that diagnostic, the
reference calculation showed that increasing one corridor from 15 to 30 stops reduced aggregate
accessibility by 18\% because added dwell time harmed through-travel. This counterexample contradicts
the assumed monotonic relation, so the diagnostic is excluded from every result here.
\FloatBarrier

\section{Complete Per-Channel Prediction Results}
\label{app:channel_results}

\begin{figure}[h]
  \centering
  \includegraphics[width=\textwidth]{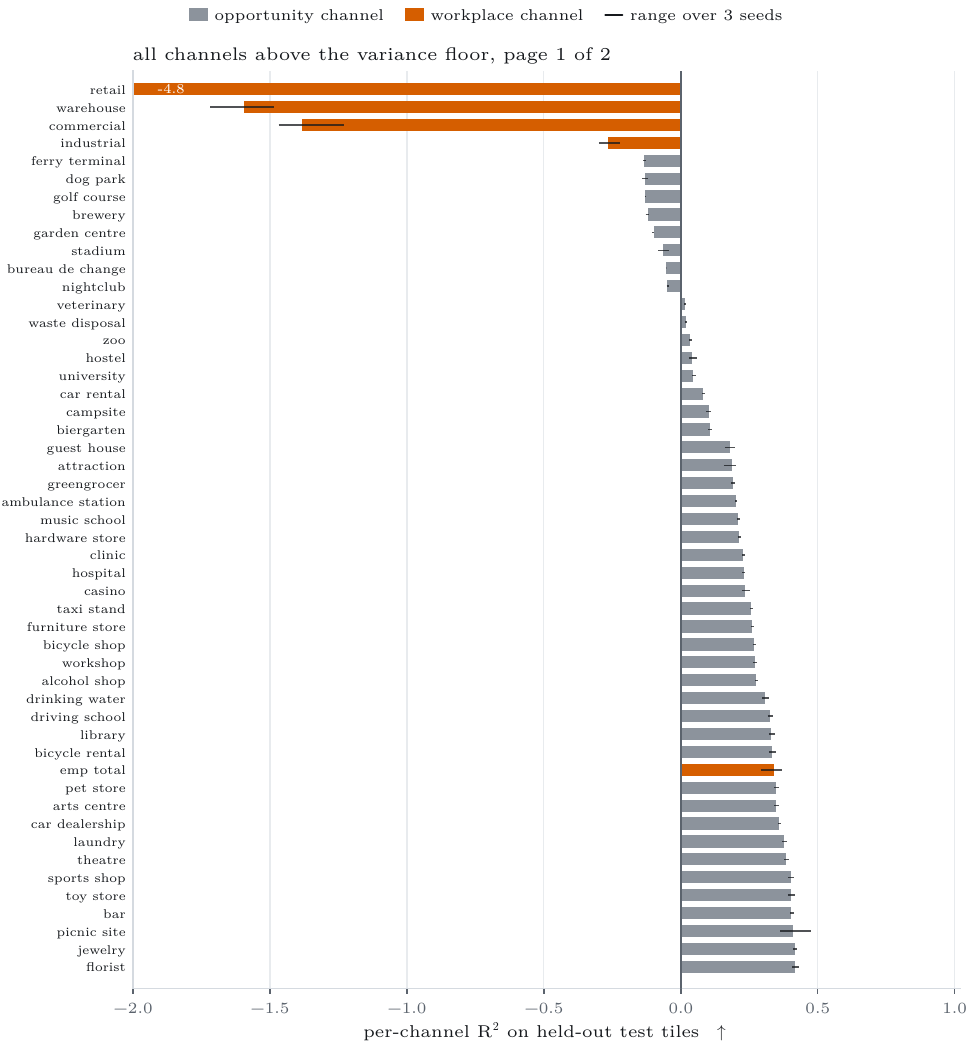}
  \caption{\textbf{Per-channel predictive agreement, lower half of the above-floor ranking.}
  Bars are means over the three width-152 \dstwo\ checkpoints; horizontal marks show the observed
  seed range. Opportunity and workplace channels are distinguished by color.}
  \label{fig:channel_full_1}
\end{figure}

\begin{figure}[h]
  \centering
  \includegraphics[width=\textwidth]{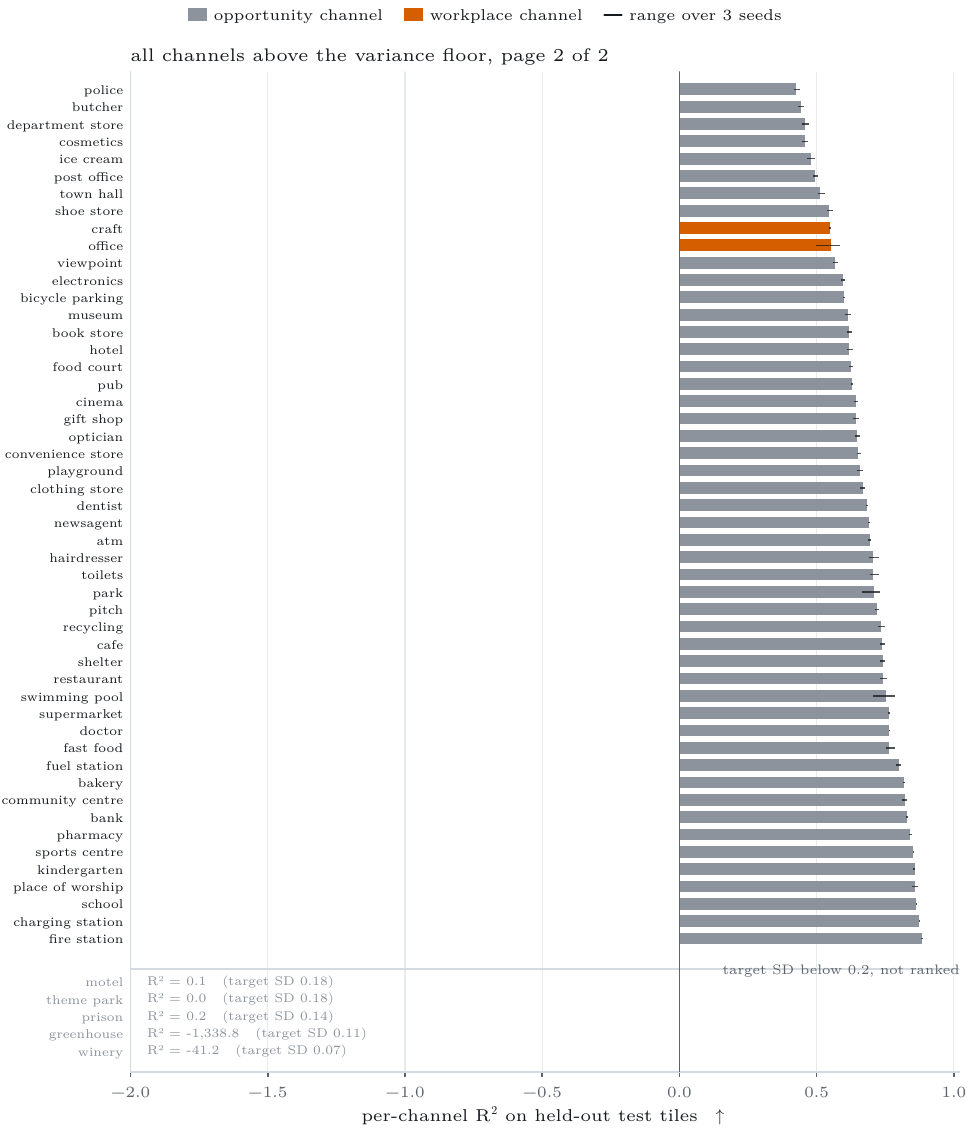}
  \caption{\textbf{Per-channel predictive agreement, upper half and unranked low-variance
  channels.} The separated rows below the rule have target SD below 0.2 and are displayed for
  completeness rather than compared by bar length.}
  \label{fig:channel_full_2}
\end{figure}

Figures~\ref{fig:channel_full_1} and \ref{fig:channel_full_2} split the full ranking across two
page-sized panels. The 100 channels above the target-SD floor are ordered from lower to higher mean
$R^2$. Five sub-floor channels are listed separately on the second page with their values and are
not assigned ranks.

\FloatBarrier

\section{Training and Geographic Split Details}
\label{app:training}

\begin{table}[h]
\centering
\small
\caption{\textbf{Shared training protocol.}}
\label{tab:training_protocol}
\begin{tabularx}{\textwidth}{@{}lX@{}}
\toprule
Setting & Value \\
\midrule
Tile geometry & $200\times200$ cells in EPSG:3857; 105 projected metres per cell \\
Input/target reduction & 32 PCA components; training-only fit \\
Loss and optimizer & Smooth L1; AdamW at $10^{-3}$ with cosine schedule \\
Training duration & at most 1,000 epochs; best validation checkpoint \\
Regularization & dropout 0.1 \\
Initialization & three seeds for each controlled configuration \\
Target normalization & fixed per-channel training statistics \\
Effective batch & 48 for large/matched models; 96 for native \ufone/\raone \\
\bottomrule
\end{tabularx}
\end{table}

Effective batch size is fixed within each architecture but differs between native-width and
matched models, confounding native-width comparisons.
A target-informed tile-mean predictor has MAE $2.760$; it uses held-out target means and is
diagnostic, not deployable.

Test membership is defined by tile centroid inside either a 35-km-radius disc centered at Munich
($11.576^\circ$E, $48.137^\circ$N) or a Mecklenburg-Vorpommern bounding box
($12.4$--$14.4^\circ$E, $53.2$--$54.5^\circ$N). Training footprints intersecting a 25.5-km
projected buffer are excluded. In EPSG:3857 coordinates, a tile spans 21 km per side and has a 14.85-km
half-diagonal, so the buffer exceeds the footprint non-overlap requirement. The final split has
14,488 training, 1,610 validation, and 818 test tiles; 81 test centroids lie in Munich and 737 in
Mecklenburg-Vorpommern. Pooled prediction metrics use all 818 tiles; regional metrics use the
corresponding subsets.

\end{document}